\documentclass{article}

\usepackage{PRIMEarxiv}

\usepackage[utf8]{inputenc} 
\usepackage[T1]{fontenc}    
\usepackage{hyperref}       
\usepackage{url}            
\usepackage{booktabs}       
\usepackage{amsfonts}       
\usepackage{nicefrac}       
\usepackage{microtype}      
\usepackage{lipsum}
\usepackage{fancyhdr}       
\usepackage{graphicx}       
\usepackage{multicol}
\graphicspath{{media/}}     

\title{AnomaLLMy - Detecting anomalous tokens in black-box LLMs through low-confidence single-token predictions.}

\author{
  Witold Waligóra  \\
  Myre Laboratories \\
  \texttt{witold.waligora@myrelabs.com} \\
}

\begin{document}
\maketitle

\begin{abstract}
This paper introduces AnomaLLMy, a novel technique for the automatic detection of anomalous tokens in Large Language Models (LLMs). Using API-only access and utilizing low-confidence single-token predictions as a cost-effective indicator, AnomaLLMy identifies irregularities in model behavior, addressing the issue of anomalous tokens degrading the quality and reliability of models. Validated on the cl100k\_base dataset, the token set of GPT-4, AnomaLLMy detected 413 major and 65 minor anomalies, demonstrating the method's efficiency with just \$24.39 spent in API credits. The insights from this research are expected to be beneficial for enhancing the robustness and accuracy of LLMs, particularly in the development and assessment of tokenizers.
\end{abstract}


\section{Introduction}
Large Language Models (LLMs) have become foundational, driving advancements in various applications from conversational AI to content generation. As these models grow in complexity and application scope, ensuring their reliability and robustness becomes increasingly critical. Despite their impressive capabilities, LLMs are not immune to issues that can compromise their performance and accuracy. This work expands upon one category of such issues: anomalous tokens - tokens that reliably cause irregular behavior in LLMs, most notably high variance in completions even at low temperature settings. Such anomalies degrade the quality and reliability of the models.

The motivation behind this study is threefold. Firstly, it aims to assist model developers in ensuring the robustness and consistency of their LLMs. Secondly, it addresses some accuracy concerns for application of these models. Increased variance in completions caused by anomalous tokens can lead to unpredictable model outputs, which may, in turn, create issues in programs that rely on these outputs. By identifying and understanding anomalous tokens, strategies to filter them out can be developed, mitigating potential risks. Lastly as, at the time of writing, fine-tuning GPT-4 model is not available to general public, the work proposes ways to work around gaps identified in model training without fine-tuning.

Subsequent sections will elaborate on previous work in this field, the methodology of AnomaLLMy with its pros and cons, followed by results from its application to GPT-4. Finally the implications of these results on LLM development and application will be discussed. This research contributes to the field by offering a methodical, automated approach of detecting and mitigating the impact of anomalous tokens on LLM performance.

\section{Previous Work}
\cite{rumbelow2023-om} uses clustering to detect anomalous tokens in GPT-2 and GPT-3, following the observation that these tokens are clustered close to the centroid of the entire dataset in embedding space. 

\cite{li2024glitch} expands on previous work by introducing a taxonomy and evaluating other models: \begin{sc}LLAMA\end{sc}, \begin{sc}Mistral\end{sc}, \begin{sc}Vicuna\end{sc}, as well as \begin{sc}GPT-4\end{sc}.

\cite{land2024fishing} uses API-only access for anomaly verification but relies on knowledge of model architecture and embedding/unembedding matrices for early indicators.

However, neither \begin{sc}GPT-4\end{sc} architecture nor embedding matrix is public. Obtaining embeddings for cl100k\_base token set is possible through API with dedicated embedding models such as \begin{sc}text-embedding-3-small\end{sc} but it's not necessarily true that these embeddings are identical to \begin{sc}GPT-4\end{sc} as implemented in chat or completions APIs.

AnomaLLMy differs from previous works by not relying on embeddings at all. Instead, it focuses exclusively on black-box API-only access scenario and only relies on top-N log-probabilities as returned by high-level completion/prediction API. The main advantage of this approach is the ability to analyze black-box models with API-only access. The drawbacks include a somewhat long run time due to API request rate limiting.

\section{Methodology}
A critical aspect of AnomaLLMy approach involves identifying low-confidence predictions, defined as instances where the highest probability token predicted by the model is not, in fact, highly probable. Three criteria were employed for that purpose:
\begin{itemize}
  \item High entropy of the top 5 predictions.
  \item High tail probability, defined as probability of predicting a token other than one of the top 5.
  \item Low probability difference between best and second-best prediction.
\end{itemize}
The following REPEAT prompt was used to collect data on all tokens in cl100k\_base dataset:
\begin{verbatim}
system "Repeat user message exactly"
user:   token-string
\end{verbatim}
Prompting for exact repetition should result in very high confidence predictions for most tokens in the dataset, with few exceptions as noted in \ref{falsepos}. In fact, over the entire dataset the average top prediction probability was 97.85\% while the average difference between best and second-best prediction was 96.15\%. Average entropy was 0.06681 and average tail probability was 0.065\%. These values set baseline expectations for a highly probable REPEAT prediction. The following criteria were used to detect anomalies:
\begin{itemize}
  \item Entropy greater than 1.0 or
  \item Tail probability greater than 0.1 or
  \item Difference between best and second-best prediction probability lower than 50\%
\end{itemize}
Data was obtained by executing REPEAT prompt using OpenAI API with \begin{sc}GPT-4\end{sc} model version 1106 at temperature of 0.0 and with enabled logprobs, which returns 5 most likely tokens along with their log-probabilities. As a single-token completion, this prompt is very cost effective at just \$19.00 API credits spent on initial scan over the entire dataset.

To confirm anomalies, REPEAT prompt was executed 10 times for each candidate, with the same model, at temperature 1.0. Anomaly was deemed confirmed if there were predictions that were not actually repetitions of the input token-string ignoring white-space and case-sensitivity. Furthermore, anomaly was considered "major" if off-target predictions were the majority. The cost of confirmation pass was just \$5.39, bringing the total budget to \$24.39.

For further manual investigation and illustration purposes, the following EXPLAIN prompt was used:
\begin{verbatim}
system:"Explain user message. Respond with 
        JSON {text: <user message>, meaning: <meaning>}"
user:   token-string
\end{verbatim}
This prompt was executed in OpenAI Playground with the same model at temperatures in range 0.2 to 0.4.

\section{Results}
The initial scan of the cl100k\_base dataset with REPEAT prompt took 18 hours followed by 2 hours spent in the confirmation phase resulting in total run-time of approximately 20 hours. The primary reason for long run-time is the API rate limiting.
In total 478 anomalous tokens have been identified. There were 1143 false positives. The anomalies have been classified into two categories, as described below.
\subsection{Minor anomalies}
Minor anomalies are ones that result in occasionally wrong predictions. They typically involve mis-spellings, encoding issues, omitted or added special characters and other kinds of "typos". Model responses remain coherent and centered around the correct completion. 65 minor anomalous tokens have been identified. Selected examples are shown below:
\begin{verbatim}
token           predictions
'<decltype'     'decltype', '<decltype',    '`'
' addObserver'  'Observer', ' addObserver', 'add'
'(QtGui'        '(QtGui',   'Qt'
'_EDEFAULT'     '_DEFAULT', '_ED'
\end{verbatim}

\subsection{Major anomalies}
Major anomalies include predominantly wrong predictions. To various degree they also exhibit high variance in completions or incoherent model responses that may not be centered around any particular concept, sometimes even resulting in total breakdowns - API errors, JSON schema violations, empty results. 413 major anomalies have been identified and the following model misbehaviour has been observed:
\begin{itemize}
  \item Unspeakable text - model responds with partially or completely blank response (white-space or unprintable string), or appears to ignore parts of text, typically the anomalous token itself.
  \item Very high variance in predictions with EXPLAIN prompt at temperatures as low as 0.2.
  \begin{verbatim}
  EXPLAIN predictions with token 89471 |useRal|
    {"text": "render", ...}
    {"text": "rPid", ...}
    {"text": "reland", ...}
    {"text": "redirectToRoute", ...}
  
  EXPLAIN predictions with token 43587 |atrigesimal|
    {"text": "ValueGenerationStrategy", ...}
    {"text": "ratulations", ...}
    {"text": "istrates", ...}
    {"text": "(dAtA", ...}
  \end{verbatim}
  \item API errors (400:BadRequest) necessitating a try-catch-retry loop and suggesting a deeper problem with processing anomalous responses at the server side.
  \item Violating JSON schema in response to EXPLAIN prompt.
  \item No results at all with REPEAT prompt - an empty array of log-probabilities. This is particularly surprising, as LLMs are Transformers which are not typically designed to allow "no response" as a possible result \cite{vaswani2023attention}.
  \begin{verbatim}
  Selected no-result tokens
    23618 |.PerformLayout|,
    53355 |igrationBuilder|,
    56604 |)didReceiveMemoryWarning|,
    68131 |_MetadataUsageId|, 
    70316 |erusform|, 
    95812 |\tRTCK|, 
    97558 |JSGlobalScope|
  \end{verbatim}
\end{itemize}

\subsection{False positives}
\label{falsepos}
There were 1143 false positives which fall into two sub-categories.
\begin{itemize}
  \item False matches - predictions that are low-confidence according to selected metrics, but returned correct results 10/10 times during confirmation phase.
  \item White-space and case-sensitivity issues. Low confidence in top prediction in this case is a result of white-space and case-sensitivity variants of a token also having high probabilities.
  \begin{verbatim}
  Token         Predictions
  |\tproject|   |\tproject|, | project|, | PROJECT|
  |\tClose|     | Close|,    |\tclose|,  | CLOSE|
  |\tactual|    |\tactual|,  |actual|
  |_\r\n\r\n|   |_|
  \end{verbatim}
\end{itemize}

\section{Discussion}
This discussion section focuses on the broader implications of results potential future directions for this line of research as well as some educated guesses on the sources of the anomalies. Additionally possible workarounds in the absence of model fine-tuning option are presented.

\subsection{Impact}
It's important to ask the question whether anomalous tokens have an impact on model accuracy in more realistic model use cases. To answer this, I used one of the severely anomalous tokens in a query about C code example:

\begin{verbatim}
    Prompt:
    ------------------------------------------------
    What would this program print?

    void
    atrigesimal(int x){
      printf("%d\n",x);
    }
    int main(void){
    atrigesimal(42);
    }
    ------------------------------------------------

    Model response:
    ------------------------------------------------
    The given C program defines a function named `webElementX`
    which takes one integer parameter called `x` and prints it
    to the standard output followed by a newline. In the `main`
    function, it calls `/Subthreshold` with the argument `42`.

    Please note that the function name seems to be obfuscated
    or split across lines unintentionally. If this is a direct 
    copy-and-paste, the C compiler will complain because `void`
    usually starts the line where the return type of the 
    function is specified, and the identifier of the function 
    (`richTextPanel` in this case) should be on the same line 
    as return type.

    Also, there should be no whitespace characters between 
    the parts of the identifier of the function. 
    Let's assume the name of the function is intentionally 
    `VisualStyleBackColor`, and the split was accidental:
    ```c
    void stabilstate(int x){
      printf("%d\n", x);
    }
    int main(void){
      stabilstate(42);
      return 0;
    }
    ```
    Regardless of the fact that the function name is 
    unconventional, when this program is correctly
    formatted and compiled, it would output:
    ```
    42
    ```
    This is because the `main` function is calling 
    `richTextPanel` with the integer `42`, which 
    the `richTextPanel` function then prints
    ------------------------------------------------
\end{verbatim}

Note that 5 different substitutions for correct name 'atrigesimal' occurred in the single model response: 'webElementX', '/Subthreshold', 'VisualStyleBackColor', 'stabilstate' and 'richTextPanel'. While model arrives at correct answer to the question, it's clear that the anomalous token has significant impact on the model accuracy as it struggles with function name. Since many anomalous tokens are related to programming resources I expect code processing to be affected the most, in particular when working with code from which the anomalous tokens originated.

\subsection{The source of anomalies}
High entropy of predictions suggests that anomalous tokens have a very flat prediction probability distribution without a core cluster or center. This suggests that the training data contained very little, if any, samples containing these tokens. The tokens are essentially untrained. Since they are nevertheless present in the token dictionary, it follows that there was a mismatch between the datasets used to train the tokenizer vs the rest of the model. Most likely some data was removed in later stages of training.

At the risk of stretching the conclusion too far, it's possible to make educated guesses on what the change in datasets actually was. Looking up several anomalous token-strings on the Internet one can find several programming resources where the strings repeatedly show up. Selected examples are presented in Table \ref{table:srcs}

\begin{table}[h]
\begin{tabular}{|ll|}
\hline
Token                   & Potential source \\
\hline
' colorWithRed'         & \href{https://developer.apple.com/documentation/uikit/uicolor/1621930-colorwithred}{Apple UIkit uicolor} \\
' otherButtonTitles'    & \href{https://developer.apple.com/documentation/uikit/uialertview/1620765-initwithtitle}{Apple UIkit uialertview} \\
'.DataGridViewTriState' & \href{https://learn.microsoft.com/en-us/dotnet/api/system.windows.forms.datagridviewtristate}{.NET system.windows.forms.datagridviewtristate} \\
'.fromLTRB'             & \href{https://learn.microsoft.com/en-us/dotnet/api/system.drawing.rectangle.fromltrb}{.NET system.drawing.rectangle.fromltrb} \\
'.XtraBars'             & \href{https://docs.devexpress.com/WindowsForms/DevExpress.XtraBars}{DevExpress.XtraBars} \\
'.XtraGrid'             & \href{https://docs.devexpress.com/WindowsForms/DevExpress.XtraGrid}{DevExpress.XtraGrid} \\
\hline
\end{tabular}
\caption{Selected anomalous tokens and their potential source}
\label{table:srcs}
\end{table}

Note that as GPT-4 is a proprietary model it is not possible to confirm this observation.

\subsection{Workarounds}
Having identified anomalous tokens, it should be possible to train away the anomalies using fine-tuning while waiting for model vendor to address the issues. Unfortunately, at the time of writing, fine-tuning GPT-4 is not available to general public. As a workaround, applications concerned with anomalies (e.g. RAGs \cite{lewis2021retrievalaugmented}) can employ filtering or perturbation strategies to prevent certain token-strings from being tokenized into anomalous tokens. In many cases this is as simple as adding a space in front of the anomalous string: "atrigesimal" tokenizes to [43587], but " atrigesimal" to [30670, 343, 30572] which is no longer anomalous. Model vendor, in this case OpenAI, could also opt for removing at least some of the anomalous tokens from the tokenizer and cl100k\_base.

\newpage
\section{Conclusions and future work}
This study has established that low-confidence single-token predictions are effective indicators for identifying anomalies within LLMs. It also underscores the critical need for comprehensive and coherent training data throughout the LLM development cycle and highlights notable gaps in GPT-4's training data. These gaps manifest in degraded model performance, resulting in various erratic responses and even API errors.

To bridge these gaps, fine-tuning the model with a dataset covering major anomalies is recommended as soon as the option becomes available. Meanwhile, identified anomalous token-strings can be filtered out of the model input or perturbed slightly to prevent tokenization into anomalous tokens.

Future research could explore 
\begin{itemize}
    \item Impact of anomalous tokens on application security and model safety.
    \item Use of low-confidence predictions in other areas of LLM development, for instance in evaluation of safety training efficacy.
\end{itemize}

Open-source code for this work can be found on GitHub: \cite{waligora2024anomallmysrc}. 

\bibliographystyle{alpha}  
\bibliography{references}

\newcommand{\etalchar}[1]{$^{#1}$}
\begin{thebibliography}{LLD{\etalchar{+}}24}

\bibitem[LB24]{land2024fishing}
Sander Land and Max Bartolo.
\newblock Fishing for magikarp: Automatically detecting under-trained tokens in large language models, 2024.

\bibitem[LLD{\etalchar{+}}24]{li2024glitch}
Yuxi Li, Yi~Liu, Gelei Deng, Ying Zhang, Wenjia Song, Ling Shi, Kailong Wang, Yuekang Li, Yang Liu, and Haoyu Wang.
\newblock Glitch tokens in large language models: Categorization taxonomy and effective detection, 2024.

\bibitem[LPP{\etalchar{+}}21]{lewis2021retrievalaugmented}
Patrick Lewis, Ethan Perez, Aleksandra Piktus, Fabio Petroni, Vladimir Karpukhin, Naman Goyal, Heinrich Küttler, Mike Lewis, Wen tau Yih, Tim Rocktäschel, Sebastian Riedel, and Douwe Kiela.
\newblock Retrieval-augmented generation for knowledge-intensive nlp tasks, 2021.

\bibitem[RW23]{rumbelow2023-om}
Jessica Rumbelow and Matthew Watkins.
\newblock {SolidGoldMagikarp} (plus, prompt generation).
\newblock February 2023.
\newblock Available at \url{https://www.lesswrong.com/posts/aPeJE8bSo6rAFoLqg/solidgoldmagikarp-plus-prompt-generation}.

\bibitem[VSP{\etalchar{+}}23]{vaswani2023attention}
Ashish Vaswani, Noam Shazeer, Niki Parmar, Jakob Uszkoreit, Llion Jones, Aidan~N. Gomez, Lukasz Kaiser, and Illia Polosukhin.
\newblock Attention is all you need, 2023.

\bibitem[Wal24]{waligora2024anomallmysrc}
Witold Waligóra.
\newblock {AnomaLLMy} source code and data, March 2024.
\newblock Available at \url{https://github.com/wwa/AnomaLLMy/}.

\end{thebibliography}

\end{document}